\documentclass[conference]{IEEEtran}

\newsavebox{\boxone}
\newsavebox{\boxtwo}
\newsavebox{\boxthree}

\newlength{\narrow}
\newlength{\cnarrow}
\newcommand{\topline}{
  \hrule
  \vskip .5\baselineskip}
\newcommand{\bottomline}{
  \vskip 2pt
  \hrule}

\newcommand{\chbox}[2]{
  \hbox to #1{\hss\vtop{#2}\hss}}

\newcommand{\nchbox}[1]{
  \chbox{\narrow}{#1}}

\newcommand{\cnchbox}[1]{
  \chbox{\cnarrow}{#1}}

\newcommand{\fcode}[1]{
  
  \chbox{\textwidth}{\tgrind\input{#1}}}

\newcommand{\ncode}[1]{
  
  \chbox{\narrow}{\tgrind\input{#1}}}

\def\nfig#1#2#3{
  \vtop{\nchbox{#1}
  \hbox to\narrow{\parbox{\narrow}{\caption{#2}\label{#3}}}}}

\newcommand{\cncode}[1]{
  \chbox{\cnarrow}{\tgrind\input{#1}}}

\def\codefiggen[#1]#2#3#4#5#6{
  \begin{figure}[#1]
  #5
  \fcode{#2}
  \center\parbox{.9\textwidth}{\caption{#3}\label{#4}}
  #6
  \end{figure}}

\def\codefig[#1]#2#3#4{
  \codefiggen[#1]{#2}{#3}{#4}{}{}}

\def\codefigline[#1]#2#3#4{
  \codefiggen[#1]{#2}{#3}{#4}{\topline}{\bottomline}}

\def\doublefiggen[#1]#2#3#4#5#6#7#8#9{
  \begin{figure}[#1]
  #8
  \hbox to \textwidth{
  \nfig{#2}{#3}{#4}
  \hfil
  \nfig{#5}{#6}{#7}}
  #9
  \end{figure}}

\def\doublefig[#1]#2#3#4#5#6#7{
  \doublefiggen[#1]{#2}{#3}{#4}{#5}{#6}{#7}{}{}}

\def\doublefigline[#1]#2#3#4#5#6#7{
  \doublefiggen[#1]{#2}{#3}{#4}{#5}{#6}{#7}{\topline}{\bottomline}}

\def\doublecodefig[#1]#2#3#4#5#6#7{
  \doublefig[#1]{\ncode{#2}}{#3}{#4}{\ncode{#5}}{#6}{#7}}

\def\doublecodefigline[#1]#2#3#4#5#6#7{
  \doublefigline[#1]{\ncode{#2}}{#3}{#4}{\ncode{#5}}{#6}{#7}}

\newcommand{\codepair}[4]{\vbox{
  \hbox{\ncode{#1} \hfil \ncode{#3}}
  \vskip .3\baselineskip plus .3\baselineskip
  \hbox{\hbox to\narrow{#2\hfil} \hfil \hbox to\narrow{#4\hfil}}}}

\def\codepairfig[#1]#2#3#4#5#6#7{
  \begin{figure}[#1]
  \codepair{#2}{#3}{#4}{#5}
  \center\parbox{.9\textwidth}{\caption{#6}}
  \label{#7}
  \end{figure}}

\def\cncodepairfiggen[#1]#2#3#4#5#6#7{
  \begin{figure}[#1]
  #6
  \hbox{\cncode{#2}\hfil\cncode{#3}}
  \center\parbox{.9\columnwidth}{\caption{#4}\label{#5}}
  #7
  \end{figure}}

\def\cncodepairfig[#1]#2#3#4#5{
  \cncodepairfiggen[#1]{#2}{#3}{#4}{#5}{}{}}

\def\cncodepairfigline[#1]#2#3#4#5{
  \cncodepairfiggen[#1]{#2}{#3}{#4}{#5}{\topline}{\bottomline}}

\def\doublefigOnecap*[#1]#2#3#4#5{
  \begin{figure*}[#1]
  \hbox to \textwidth{
  \nchbox{#2}
  \hfil
  \nchbox{#3}}
  \caption{#4}
  \label{#5}
  \end{figure*}}

\def\doublefigOnecap[#1]#2#3#4#5{
  \begin{figure}[#1]
  \topline
  \hbox to \columnwidth{
  \cnchbox{#2}
  \hfil
  \cnchbox{#3}}
  \caption{#4}
  \label{#5}
  \bottomline
  \end{figure}}

\def\PSfig[#1]#2#3#4{
 \begin{figure}
 \centerline{\psfig{file=#2,width=\columnwidth}}
 \caption{{#3}} 
 \label{#4}
 \end{figure}}

\def\PSfiglines[#1]#2#3#4{
 \begin{figure}[#1]
 \topline
 \centerline{\psfig{file=#2,width=\columnwidth}}
 \caption{{#3}} 
 \label{#4}
 \bottomline
 \end{figure}}

\def\PSfiglinesht[#1]#2#3#4#5{
 \begin{figure}[#1]
 \topline
 \centerline{\psfig{file=#2,height=#3}}
 \caption{{#4}} 
 \label{#5}
 \bottomline
 \end{figure}}

\def\doublePSfig[#1]#2#3#4#5#6{
  \doublefigOnecap[#1]
    {\cnchbox{\psfig{file=#2,height=#4}}}
    {\cnchbox{\psfig{file=#3,height=#4}}}
    {#5}
    {#6}}

\newlength{\boxwidth}
\def\tabdoublecode#1#2#3#4{
 \begin{figure*}[t]
 \topline\vs{-.4}
 \hbox to \columnwidth{
 \vtop{\small
 \begin{tabbing}
 #1
 \end{tabbing}}
 \hfil
 \hfil
 \hfil
 \vtop{\small
 \begin{tabbing}
 #2
 \end{tabbing}}
 }
 \caption{#3\label{#4}}
 \bottomline
 \end{figure*}
}
\def\tabtriplecode#1#2#3#4#5{
 \begin{figure}
 \topline\vs{-.4}
 \hbox to \columnwidth{
 \vtop{\small
 \begin{tabbing}
 #1
 \end{tabbing}}
 \hfil
 \hfil
 \hfil
 \vtop{\small
 \begin{tabbing}
 #2
 \end{tabbing}}
 \hfil
 \hfil
 \hfil
 \vtop{\small
 \begin{tabbing}
 #3
 \end{tabbing}}
 }
 \caption{#4\label{#5}}
 \bottomline
 \end{figure}
}

\newcommand{\eg}{{\em e.g.}}

\newcommand{\ie}{{\em i.e.}}

\newcommand{\vs}[1]{\vspace{#1cm}}
\newcommand{\be}{\begin{equation}}
\newcommand{\ee}{\end{equation}}

\newcommand{\bdesc}{\begin{description}}
\newcommand{\edesc}{\end{description}}
\newcommand{\benum}{\begin{enumerate}}
\newcommand{\eenum}{\end{enumerate}}
\newcommand{\bitem}{\begin{itemize}}
\newcommand{\eitem}{\end{itemize}}
\newcommand{\bcenter}{\begin{center}}
\newcommand{\ecenter}{\end{center}}
\newcommand{\btabular}{\begin{tabular}}
\newcommand{\etabular}{\end{tabular}}
\newcommand{\beqnarr}{
 \begin{eqnarray}}
\newcommand{\eeqnarr}{\end{eqnarray}}

\long\def\symbolfootnote[#1]#2{\begingroup
\def\thefootnote{\fnsymbol{footnote}}\footnote[#1]{#2}\endgroup}

\newcommand{\eat}[1]{}

\usepackage{algorithmic}
\usepackage{amssymb, amsmath}
\usepackage{fixltx2e}
\usepackage{stfloats}
\usepackage{url}
\usepackage{graphicx}
\usepackage{epstopdf}
\usepackage{cite}
\usepackage[colorinlistoftodos]{todonotes}
\usepackage{soul}
\usepackage{float}

\newcommand{\superscript}[1]{\ensuremath{^{\textrm{#1}}}}

\begin{document}
\title{Detecting Suicidal Ideation in Chinese Microblogs with Psychological Lexicons}

\author{\IEEEauthorblockN{Xiaolei Huang\superscript{1,2} ~~~
Lei Zhang\superscript{1} ~~~
Tianli Liu\superscript{3} ~~~
David Chiu\superscript{4} ~~~
Xin Li\superscript{*2} ~~~
Tingshao Zhu\superscript{*1,5}
}
\IEEEauthorblockA{\superscript{1} Institute of Psychology, Chinese Academy of Sciences(CAS), China}
\IEEEauthorblockA{\superscript{2} China Networking Information Center, CAS, China}
\IEEEauthorblockA{\superscript{3} Institute of Population Research, Peking University, China}
\IEEEauthorblockA{\superscript{4} Department of Mathematics and Computer Science, University of Puget Sound, USA}
\IEEEauthorblockA{\superscript{5} Key Lab of Intelligent Information Processing, Institute of Computing Technology, CAS, China}
}
\maketitle

\begin{abstract}
Suicide is among the leading causes of death in China. However,
technical approaches toward preventing suicide are challenging and remaining under development.
Recently, several actual suicidal cases were preceded by users who posted microblogs with suicidal ideation to \textit{Sina Weibo},
a Chinese social media network akin to \textit{Twitter}.
It would therefore be desirable to detect suicidal ideations from microblogs in real-time,
and immediately alert appropriate support groups, which may lead to successful prevention.
% , and we have collected this data using the \textit{Sina Weibo} API.
In this paper, we propose a real-time suicidal ideation detection system deployed over \textit{Weibo},
using machine learning and known psychological techniques.
Currently, we have identified 53 known suicidal cases who posted suicide notes on \textit{Weibo} prior to their deaths.
% using a psychological lexicon dictionary.
%We describe our approach to detect suicidal posts on a new and unique real data set.
We explore linguistic features of these known cases using a psychological lexicon dictionary, and train an effective suicidal \textit{Weibo} post detection model. $6714$ tagged posts and several classifiers are used to verify the model. By combining both machine learning and psychological knowledge,
SVM classifier has the best performance of different classifiers,
yielding an F-measure of $68.3\%$, a Precision of $78.9\%$, and a Recall of $60.3\%$.
%According to our research statistics,
%we also use those data to verify some psychological theory in a larger dataset.
\end{abstract}

% \IEEEpeerreviewmaketitle

\section{Introduction}
The World Health Organization (WHO) recently reported that $800,000$ suicidal deaths occurred worldwide in 2012~\cite{whosuicide},
which translates to a person committing suicide every 40 seconds.
This unfortunate report places suicide among the leading causes of death worldwide,
particularly for young people ages 15 to 29.
% Specifically, that is an annualized global age-standardized suicide rate of 11.4 per 100,000 population, which means that a person commits suicides every  40 seconds.
In recent years, the suicide problem in China has drawn much attention.
Former research~\cite{michael2002suicide} reported that around $290,000$ Chinese people died due to suicide every year,
accounting for roughly $36\%$ of all suicidal cases worldwide.
Although the number has decreased during to past several years,
WHO~\cite{whosuicide} reported that there were still $120,730$ Chinese people committed suicide in 2012.
% Nearly 2 million Chinese people have suicide attempts each year.

For years, psychologists have been working on suicidal prevention,
they found that suicidal ideation can be clearly identified from victims' writings and posts.
Wong, \textit{et al.} analyzed suicide notes obtained from the coroner's records in Hong Kong~\cite{wong2009suicide}.
In this study,
they found several characteristics separating suicidal from non-suicidal people and indicated reasons for the increasing suicide rates in Hong Kong,
including media and social changes.
However, because this data is only obtainable from key players in clinical investigations, the volume of pre-suicidal data is very limited.
% but with a lack of information technology,
% they can not carry out the analysis in time and large scale.

Now in the era of Web 2.0, a multitude of Social Network Sites (SNS), \eg, Twitter and Facebook,
have emerged as essential platforms for communicating with others in real-time.
Users often express their emotions and opinions by posts on SNS,
which becomes their preferable choices. Jashinsky, \textit{et al.} reported suicidal behaviors in Twitter,
they pointed out that the association between social media and suicide has become a public health concern~\cite{jashinsky2013tracking}.
%Likewise in China, many users have expressed their emotions publicly on \textit{Sina Weibo}, known as the ``Chinese Twitter.''

Due to various reasons,
people may not have an appropriate way to alleviate their depressions in real life.
Previous study~\cite{finn2006mmpi} indicated that some people preferred to understate symptoms to avoid negative consequences,
or some people lacked the self awareness to report to others,
in addition, they might be lack of trust in others.
In several real cases,
users chose to post microblog notes to describe their suicidal ideation on \textit{Weibo}.
However, these suicidal behaviors on \textit{Weibo} were not detected, or had simply not raised enough attention. We believe that early detection of suicidal behaviors on SNS may help prevent deaths.

In this paper, we build a suicide ideation detection framework to analyze microblogs with various features, and explore the possibility of monitoring suicidal behaviors. The current suicidal detection initiatives are very labor-intensive and ineffective, because they rely heavily on simple keyword matching and questionnaire. In collaboration with psychologists, we leverage advanced text analysis and mining integrated with psychological techniques to build an efficient suicidal ideation detection system over SNS.
With Natural Language Processing (NLP) techniques, we propose to use text analysis techniques to study latent emotions of users, and then we design, develop, and experiment a system that can identify suicidal posts in real-time.
%We have verified every suicidal user in our experiment.

Our research has three main contributions, highlighted below.
\begin{itemize}
\item First, to the best of our knowledge, there is a lack of formal suicidal research on Chinese social media.
Our research is a unique and exploratory research on real data set.
We have verified 53 suicidal users on \textit{Weibo} post-mortem and collected over $30,000$ of their posts.
% Further, more users' verification is under working.

\item Second, we explore NLP techniques with psychological knowledge to find an approach to predict suicidal posts and achieved F-measure of $68.3\%$.
We also propose an approach to handle imbalanced data between suicidal posts and non-suicidal posts.
\item Third, interdisciplinary research between machine learning and psychology is nontrivial.
We use machine learning skills to verify and prove psychological hypothesises.
\end{itemize}

The remainder of this paper is organized as follows. Section~\ref{sec:relatedwork} presents the related work.  The background and description of our multi-hierarchical model will be discussed in Section~\ref{sec:method}. Section~\ref{sec:experiment} presents the experimental design and test results of our proposed system. Finally, future work directions and conclusion will be discussed in Section~\ref{sec:final}.

\section{Related Works}
\label{sec:relatedwork}
In this section, we give an overview on suicide detection in psychology, and then we briefly introduce current research on suicidal ideation detection.

Suicide has been intensively studied in the field of psychology. Traditional research methods mainly rely on questionnaires. Chen, \textit{et al.}, reported that ``Media Propagation'' had significant effects on suicidal behaviors~\cite{chen2013impact}. They found that simply due to seeing news reports of charcoal burning suicides, the rate of charcoal burning suicide increased.
Li, \textit{et al.}, studied one 13-year-old youth's 193 blog entries by using the Chinese Linguistic Inquiry and Word Count (CLIWC),
and found several key features in his suicidal notes, including posting frequency,
progressive self-referencing, and positive to negative emotion words ratio~\cite{li2014temporal}.

In another study, Delgado-Gomez, \textit{et al.}, acquired 849 cases from Spain hospital and scored them manually by psychological standards or questionnaires~\cite{delgado2011improving}. They used these scores to evaluate four classifiers, including Boosting, Linear Discriminate Analysis, Fisher Linear Discriminate Analysis and Support Vector Machine (SVM). They found that the score, which was analyzed manually by doctors,
could be used to identify suicidal person using classifiers.

Opinion mining and sentiment analysis techniques have also been applied in various fields~\cite{pang2008opinion}, and increasing numbers of psychologists use social media search engines to find interesting patterns in suicidal entries, such as sentiment phrases, geographic location, \textit{etc}.
Vincent, \textit{et al.}, explored the relationship between suicide and sexual orientation~\cite{silenzio2009connecting}. Jared, \textit{et al.}, tracked data from Twitter to find out suicide risk difference between states~\cite{jashinsky2013tracking}.

Applying machine learning techniques in suicide identification has gained traction in recent years.
Linguistic Inquiry and Word Count Version 2007 (LIWC2007) evaluates various emotional, cognitive,
and structural English words or phrases presented in individuals' verbal and written sentences~\cite{chung2007psychological}.
LIWC could be used to identify a post's emotional tendency.
Ramirez-Esparza, \textit{et al.},
worked on linguistic markers and themes of depression discussion by collecting information both in depressed and non-depressed individuals from legacy Internet forums using bulletin board systems (BBS)~\cite{ramirez2008psychology}.
They analyzed the text with the help of LIWC,
and they found that online depressed writers used more first person singular pronouns but less first-person plural pronouns in both English and Spanish forums.
They also found that depressed people who wrote in Spanish were more likely to relational concerns;
whereas depressed people who wrote in English were more likely to mention medical concerns.

Huang, \textit{et al.}, used basic keywords matching (similar to regular expressions) over \texttt{Myspace.com}
% ``LIKE'' in SQL,
to discover whether users have an intent to commit suicide~\cite{huang2007hunting}.
Their method  is simple and  recognized 14\% of the known suicidal users.
Li, \textit{et al.}, later proposed a hybrid method~\cite{li2012hybrid}. They noted importance of applying machine learning skills into emotional distress detection, like SVM.

Wang, \textit{et al.},  worked on classifying English clinical suicide notes into 16 different emotional categories, such as blame, fear, love, \textit{etc.}~\cite{wang2012discovering}. Their data is based on the fifth I2B2 (Informatics for Integrating Biology and the Bedside) challenge~\cite{pestian2012sentiment}, and they used SVM classifier to test different features,
with the result of F-measure around $50.04\%$, and showed the possibility for applying machine learning into predicting suicide notes.
Liakata, \textit{et al.} worked on the same data. They used a hybrid classifier,
which includes both SVM classifier and CRF classifier, with the best F-measure $45.64\%$~\cite{liakata2012three}. Pestian, \textit{et al.},
applied NLP techniques on 66 suicide notes and selected reading scores, words, concepts and other 63 features~\cite{pestian2010suicide}.
They used over ten different methods in Weka~\cite{hall2009weka},
an useful machine learning tool to classify their data, and they achieved an impressive precision of $74.4\%$,
compared to $51.0\%$ from psychiatric physician trainees and $60.9\%$ from mental health professionals.

Wang, \textit{et al.}, proposed a Chinese depression detection system~\cite{wang2013depression}.
They used three classifiers from Weka,
and their analysis and experiment data included one verified suicidal user's data, which contained 215 posts.
Their research could achieve their best F-measure of $85\%$.
\section{Methods}
\label{sec:method}
In this section, we present the background and detailed description of our system and the algorithms involved.

\begin{figure*}[htp!]
  \centering
  \includegraphics[width=0.8\textwidth]{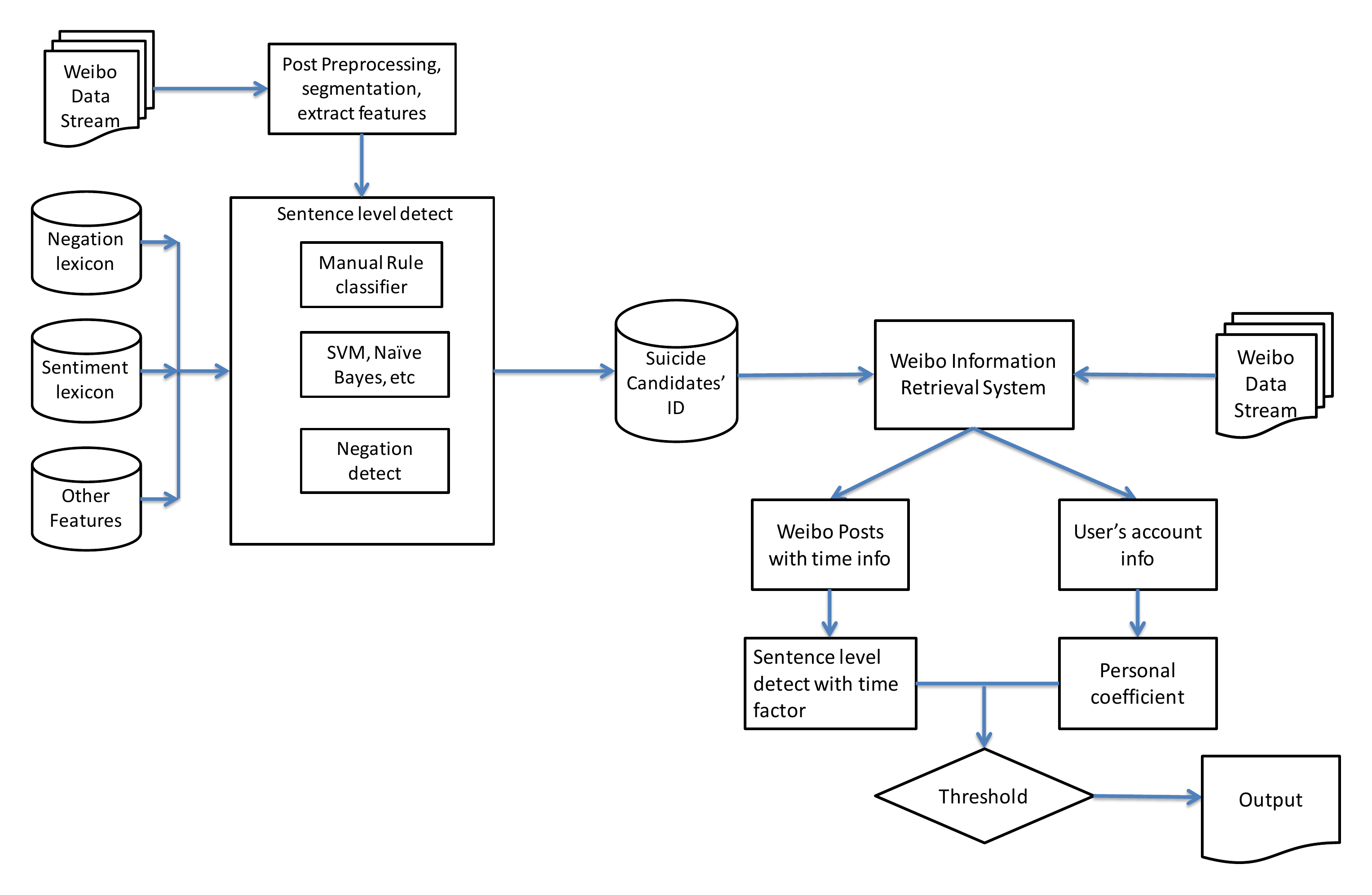}
  \caption{\label{fig:framework}Real-Time Suidical Detection Framework}
\end{figure*}

An overview of our three-level suicide detection framework is shown in Fig.~\ref{fig:framework}.
Posts collected from our crawler are used to extract several basic attributes:
1) text contents; 2) user's behaviors, such as posting time, mentions, posting type;
3) social interactions, such as the users who comment or like this post.
With the help of linguistic dictionaries,
we are able to extract a set of features from those basic attributes as the first step,
such as keywords, negations, negative emotions, and so on.
Then, we take a sample and assign weights to process the data to solve imbalanced data.
Imbalanced data refers to any data set that exhibits an unequal distribution between its classes\cite{he2009learning}.
In our data set, none suicidal posts are much more than suicidal posts.
Finally, classifiers are applied to learn the suicidal features and build a suicidal detection model.

\subsection{Data Collection and Tagging}
Our data is collected through Sina Microblog Open Platform API~\cite{sinaweibo} and our Java-based crawler.
We downloaded 53 verified suicidal users' profiles and over $30,000$ posts.
In addition, we also collected $600,000$ posts from $1,000$ thousand random non-suicidal users.
We curated  all suicidal users' posts and obtained $614$ true suicidal posts. To perform a 90-10 test,
we randomly sampled $6,140$ posts from the set of  non-suicidal users, for a total of $6,754$ posts.
After filtering some blank posts, we obtain $6,704$ posts.
% Our following research is based on these data.
%The data composition is showed in Fig. \ref{fig:composition}.
%\begin{figure}
%\centering
%\includegraphics[width=0.25\textwidth]{./pic/suicidaVslNone.jpg}
%\caption{Data Composition}
%\label{fig:composition}
%\end{figure}

\subsection{Lexicon Construction}
Because no lexicons specifically for suicidal word or phrases are available,
we extend our psychological lexicons with help of the latest version of HowNet~\cite{dong2003hownet},
a Chinese-based emotional-words resource for sentiment analysis.
Psychological dictionaries contain some depressed words or phrases often used in cases of suicide notes.
We use those dictionaries to detect and count the number of positive, negative, neutral words or phrases,
and the ratio of positive to negative words.
Examples of Chinese words and phrases in HowNet are depicted in Table~\ref{tab:hownet} below,
and examples of psychological dictionaries are shown in Table~\ref{tab:PsychologicalDictionary}.

\begin{table}[H]
\caption{Emotional words in HowNet}
\label{tab:hownet}
\centering
\begin{tabular}{c|c|c}
\hline
Emotion Words& Number& Example\\
\hline
Negative& 4370 &bad, sad, tragedy, coward\\
\hline
Positive& 4566 &happy, safe, amiable, calm\\
\hline
\end{tabular}
\end{table}

\begin{table}[H]
\caption{Psychological Dictionary}
\label{tab:PsychologicalDictionary}
\centering
\begin{tabular}{c|c|c}
\hline
Type& Number& Example\\
\hline
Terms& 3453 &dead, psychologist, insomnia, stilnox\\
&&depression, estazolam, schizophrenia \\
\hline
\end{tabular}
\end{table}

\subsection{Feature Extraction and Modeling}
\label{subsec:features}
We use Ansj~\cite{AnsjSegmentation}, a Chinese word segmentation tool,
to segment words and remove punctuation and stop words.
We choose three basic $N$-gram features: Unigram, Bigram and Trigram.
We use emotional words from lexicon to find out its emotional polarity in the sentence,
and we also compute the ratio of negative to positive emotion words.

Li, \textit{et al.}, found that suicidal victims tend to write more about themselves and less about others in posting entries,
such as I, me, myself, we, \textit{etc.}~\cite{li2014temporal}. During our research,
we found that victims also like to mention their parents or grandparents, brothers or sisters,
or even other suicidal victims. The statistics of references is shown in Table~\ref{tab:personalreference}.

\begin{table}[H]
\caption{Personal Reference Statistics}
\label{tab:personalreference}
\centering
\begin{tabular}{c|c|c|c}
\hline
Type& Suicide Ratio & None Suicide Ratio & Example\\
\hline
Self-references & 68\% & 35\% &  I, myself, me\\
\hline
Other-references & 28\% & 5\% & Dad, mother, grandpa\\
\hline
\end{tabular}
\end{table}

Syntactic features contains dependency relation, Part of Speech (POS) tagging and sentence tense.
We use Stanford Parser Core NLP~\cite{688957} to obtain the statistics of POS tag for each sentence.
Previous research~\cite{pestian2010suicide} showed suicide notes have some distinct POS tag categories.
Because the sentences on web are usually complex and ambiguous, we only obtained three types of POS tags by counting the numbers of words in the following: adjective, noun, verb.

\begin{figure}[H]
\centering
\includegraphics[scale=0.35]{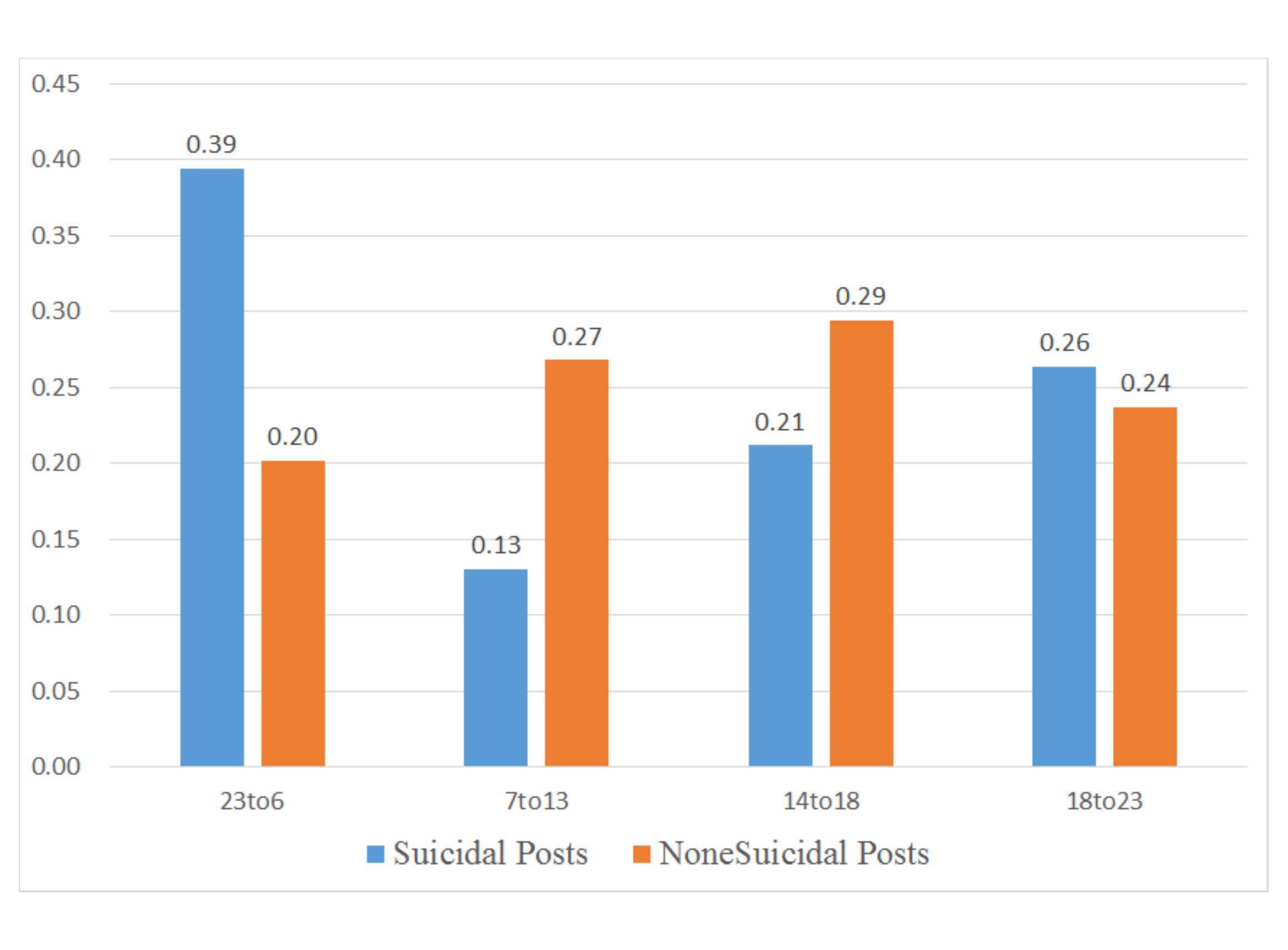}
\caption{Posting Time Comparison}
\label{fig:time}
\end{figure}

We also found that temporal features matter.
In our data, nearly $40\%$ of suicidal posts are posted at night or early in the morning from 23:00 to 06:00 in the morning,
whereas only $13\%$ of non-suicidal users posted during this period.
A plausible explanation might be some suicidal users suffer from insomnia and post about hypnotic pills,
such as tranquillizer or stilnox.
In contrast, only $13\%$ suicidal posts are posted in the morning from 07:00 to 13:00,
and over $28\%$ of non-suicidal posts are posted during this same period.
Thus, we separate 24 hours into four time periods: 23:00 to 06:00, 07:00 to 13:00, 14:00 to 18:00,
and 18:00 to 23:00. The more specific information are shown in Fig. ~\ref{fig:time}.

\begin{figure}[H]
\centering
\includegraphics[scale=0.38]{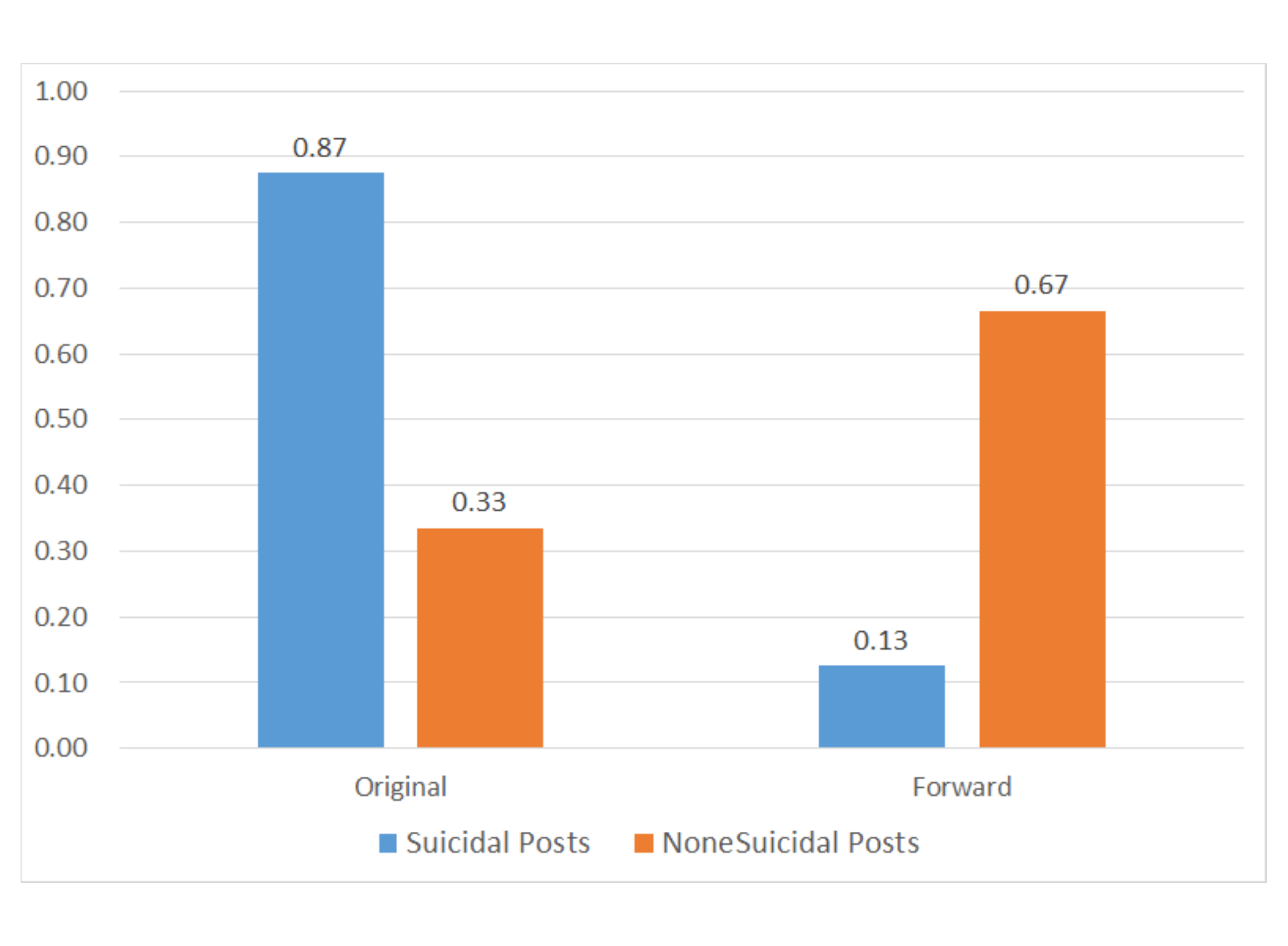}
\caption{Post's Type Comparison}
\label{fig:postType}
\end{figure}

Posting type refers to \textit{original creation} or \textit{forward posts}, which is another prominent factor.
Although forwarding others'  posts (\ie, retweet or propagation) has some effects on users' behaviors,
it appears in our data that suicidal users prefer to post original thoughts.
In our data, original posts account for $87\%$ of tagged suicidal posts,
and in contrast, original creation only accounts for $33\%$ of  non-suicidal users' posts.
The more specific comparisons are shown in Fig.~\ref{fig:postType}.

Another feature is posting frequency as time varies, which may be caused by the users' emotional variations. Research shows that the deceased posted less when they were happier~\cite{li2014temporal}.
In addition, depressed users may post \textit{longer} emotional posts than people who commit suicide impulsively, and this phenomenon is also reflected in our own research.
Thus, for every post in our data set, we compute its cycle's frequency, and then compare it with its users' average posting frequency.

Since the data is skewed,
% Previous works do not consider this problem, however suicidal data in real-world may be more skewed.
we use over-sampling to solve this problem,
and design a function to compute and change automatically the initial value of each feature.
We consider the basic weight computation in three aspects.
\begin{itemize}
\item First, the ratio of suicidal posts to none suicidal posts is chosen for simulating real-world data sets.
We denote the number of suicidal posts as $N_{s}$ and the number of non-suicidal posts as $N_{n}$.
\item Second, the occurrence of each word or other features,
such as occurrence of original created posts and forward posts,
reflect their significance in posts.
In contrast to \textit{TF-IDF}~\cite{salton1983introduction} used in information retrieval,
more frequent occurrences may be more significant, if the words are self-references or tagged suicidal.
We consider its occurrence ratio between suicidal posts to non-suicidal posts here.
We denote occurrence in each suicidal post as $Q_{s}$ and occurrence in each none suicidal post as $Q_{n}$.
\item Third, we consider several properties of lexicon dictionaries.
We consider two properties: the polarity of dictionaries, such as positive or negative and its lexicon size.
\end{itemize}

The final feature's initial weight value $W_{o}$, such as one or zero, should multiply an additional weight function, which is aim to solve skewed data problem:
\begin{gather}
\displaystyle
weight = W_o \times \log\left[\frac{\displaystyle\sum_{i=1}^{N_s}{Q_s}}{\displaystyle\sum_{i=1}^{N_s}{Q_n+1}}\right]\times\frac{N_n}{N_s}
\end{gather}
\section{Experimental Evaluation}
\label{sec:experiment}
In this section, we describe our experiment approach and discuss the results.
Our experiment's classification results are evaluated by ``Precision,'' ``Recall,'' ``F-measure,'', and ``Accuracy''.
We take whether the classifier could correctly detect the suicidal posts for ``Precision,'' ``Recall,'' and ``F-measure,'' respectively.
``Accuracy'' refers to all predictions match their labels regardless whether it is suicidal posts or not. We use 10-fold-cross validation to ensure proper separation of our $6704$ experimental data entries, and use different classifiers to perform classification.

We use the SVM classifier from LibSVM~\cite{chang2011libsvm} package for C and Java. The classifier in our experiments used a Radial Basis Function (RBF) kernel. For parameters, we use grid search with $C=16$ and $\gamma=3.0517578125e-5$. We choose several classifiers from Weka~\cite{hall2009weka}: NaiveBayes, LogisticRegression, J48 Classifier, RandomForest and SMO. All classifiers were run with default parameter values, and the results are presented in Table~\ref{tab:results}.

\begin{table}[H]
\caption{Cross-validation of performance on different classifiers}
\label{tab:results}
\centering
\begin{tabular}{|c|c|c|c|c|}
\hline
 & F-measure & Precision & Recall& Accuracy \\
\hline
 NaiveBayes &  $40.2\%$  & $\textbf{79.0\%}$ & $26.9\%$ & $92.6313\%$ \\
\hline
 LogisticRegression & 39.3\% & 55.8\% & 26.9\% & 88.0072\%\\
\hline
 J48 & 46.6\% & 68.9\% & 35.2\% & 92.5865\%\\
\hline
RandomForest & 37.4\% & 74.0\% & 25.0\% & 92.3031\%\\
\hline
 SMO & 56.8\% & 68.7\% & 48.4\% & 93.2279\%\\
\hline
 SVM & \textbf{68.3\%} & 78.9\% & \textbf{60.3\%} & \textbf{94.0036\%}\\
\hline
\end{tabular}
\end{table}

Table~\ref{tab:results} presents the performance of each classifier.
Clearly, we can see from Table~\ref{tab:results} that
the SVM classifier achieves the best ``F-measure'' with $68.3\%$, ``Recall'' with $60.3\%$, and an ``Accuracy'' over $94\%$.
If we take all posts into consideration, the ``Accuracy'' reaches over $94\%$, which means the possibility that this suicide detection system can assist psychologists in identifying true suicidal posts.
It will be a great help for suicide prevention.

However, we find that $40\%$ of suicidal posts are not identified correctly. This may mainly be due to the complexity and ambiguity of Chinese sentences on the Internet. In addition, if we extracted more features and applied more advanced text mining methods, we might achieve better results, which will be in our future work.

%\begin{table*}[htbp]\large
%\caption{Cross-validation of performance on SVM Classifier with different feature combinations}
%\label{tab:featuresresults}
%\centering
%\begin{tabular}{|c|c|c|c|c|}
%\hline
% Feature Combinations & F-measure & Precision & Recall& Accuracy \\
%\hline
% Unigram+Lexicon & 27.0\% & 79.0\% & 15.8\% & 91.8407\%\\
%\hline
% Unigram+Lexicon+Bigrams & 39.3\% & 55.8\% & 26.9\% & 88.0072\%\\
%\hline
% N-grams+Lexicons & 46.6\% & 68.9\% & 35.2\% & 92.5865\%\\
%\hline
% All+OtherFeatures & 68.3\% & 78.9\% & 60.3\% & 94.0036\%\\
%\hline
%\end{tabular}
%\end{table*}
%\par We also use SVM classifier to test different feature combinations with SVM. The results are shown in Table \ref{tab:featuresresults}. As in Table \ref{tab:featuresresults}, we can find that using single feature like lexicons, it can not achieve good results. And we can find that posting time or other syntactic features improve classification slightly.

From the feature selection Section~\ref{subsec:features}, we can find the preferences of people with high risk of suicide.
\begin{itemize}
 \item Suicidal users prefer to post their suicidal posts at night.
 \item Suicidal users prefer to post suicidal posts originally instead of forwarding others' posts, though others' emotions will propagate and affect them.
 \item Suicidal users prefer to mention themselves more frequently than none suicidal users.
\end{itemize}

This research has a number of potential applications. The trained model can be used to build an alert system, and the system is urgent for preventing and intervening suicidal users. If this system is effective in detecting suicidal posts from social networks, it may provide psychologists with advanced decision support.

\section{Conclusion \& Future Work}
\label{sec:final}
In this paper, we build a suicidal ideation detection system using psychological lexicons. Based on psychological lexicons,
we use machine learning techniques to detect suicide ideation in posts from social networks,
which is a new approach to investigate people's online behavior.
Since the social networks provide people with a new way of communication, to let them express their emotion,
which may include posts with suicide ideation,
it is very helpful to build a monitoring system that could help psychologists analyze those posts,
to identify psychological characteristics, such as suicide.
In our work, the precision of suicide ideation is $79\%$, and overall the accuracy can reach $94\%$.
The purpose of this research is to draw people's attention to problem of suicide and to show that how useful it is,
 if we could use both machine learning and psychology knowledge for identifying suicide ideation.

Our future work is undertaken in three directions: ``Topic Model'', ``Latent Social Relationship'', and ``Deep Learning''.
Topic Model~\cite{blei2003latent} is a very useful tool to analyze text.
It may find latent semantics in suicide posts, and it also can find dependency on lexicons.
We also plan to use latent social network relationships to identify suicide individuals and groups.
People who have interactions in their posts might propagate information to others.
Some research have been done on this topic~\cite{chen2013impact},
and shed lights on this part in media propagation.
Deep learning is a very popular machine learning technique in recent years.
Applying deep learning techniques in detecting suicide post text may help us build more efficient features to improve the performance of prediction.

\section*{Acknowledgments}
The authors gratefully acknowledge the generous support from National High-tech R\&D Program of China (2013AA01A606),
National Basic Research Program of China(973 Program£¬ 2014CB744600),
Key Research Program of CAS (KJZD-EW-L04) and Strategic Priority Research Program (XDA06030800) from Chinese Academy of Sciences.
We would like to thank Hongbo Chen and Xiongcai Luo for insightful discussions and suggestions,
and special thank Google Summer of Code 2014 and Portland State University for sponsoring the open-source development of this project.

\bibliographystyle{IEEEtran}
\bibliography{reference}
\end{document}